\documentclass[10pt,twocolumn,letterpaper]{article}

\usepackage{cvpr}
\usepackage{times}
\usepackage{epsfig}
\usepackage{graphicx}
\usepackage{amsmath}
\usepackage{amssymb}
\usepackage{multirow}

\usepackage[pagebackref=true,breaklinks=true,letterpaper=true,colorlinks,bookmarks=false]{hyperref}

 \cvprfinalcopy 

\def\cvprPaperID{2779} 

\ifcvprfinal\fi
\begin{document}

\title{Effective Domain Knowledge Transfer with Soft Fine-tuning}

\author{Zhichen Zhao, Bowen Zhang, Yuning Jiang, Li Xu, Lei Li, Wei-Ying Ma\\
Bytedance Inc\\
{\tt\small \{zhaozhichen.water,zhangbowen.berwyn,jiangyuning,xuli.rd,lileilab,maweiying\}@bytedance.com}
}

\maketitle

\begin{abstract}
Convolutional neural networks require numerous data for training. Considering the difficulties in data collection and labeling in some specific tasks, existing approaches generally use models pre-trained on a large source domain (\eg ImageNet), and then fine-tune them on these tasks. However, the datasets from source domain are simply discarded in the fine-tuning process. We argue that the source datasets could be better utilized and benefit fine-tuning. This paper firstly introduces the concept of general discrimination to describe ability of a network to distinguish untrained patterns, and then experimentally demonstrates that general discrimination could potentially enhance the total discrimination ability on target domain. Furthermore, we propose a novel and light-weighted method, namely soft fine-tuning. Unlike traditional fine-tuning which directly replaces optimization objective by a loss function on the target domain, soft fine-tuning effectively keeps general discrimination by holding the previous loss and removes it softly. By doing so, soft fine-tuning improves the robustness of the network to data bias, and meanwhile accelerates the convergence. We evaluate our approach on several visual recognition tasks. Extensive experimental results support that soft fine-tuning provides consistent improvement on all evaluated tasks, and outperforms the state-of-the-art significantly. Codes will be made available to the public.
\end{abstract}

\section{Introduction}
\label{sec:intro}
Convolutional neural networks (CNNs) have achieved great success on visual recognition tasks~\cite{alexnet,vgg,resnet}. While in general there is a consensus that large-scale labeled datasets are needed to train CNNs with millions of learnable parameters, for some specific tasks, \eg fine-grained categorization and infrared face recognition which need expert-level labeling~\cite{parteccv,nat,DLA} or special imaging equipments~\cite{oulu,casia}, the training datasets are difficult to scale. To boost the recognition performance of CNNs on the specific tasks, existing approaches~\cite{rcnn,contextualaction} overwhelmingly adopt a transfer learning method, namely {\it fine-tuning}: instead of training a CNN model from scratch, they utilize a CNN model pre-trained on a large-scale source image dataset such as ImageNet~\cite{imagenet} and Places~\cite{place}, then re-initialize the last classifier layer of the model with random weights and train it on the small-scale target dataset. The fine-tuning technique is simple but effective, and it has been widely used in various tasks and reaches the state-of-the-art results~\cite{DLA,nat}.

\begin{figure}[t]
\centering
\includegraphics[width=0.49\textwidth,height=0.29\textwidth]{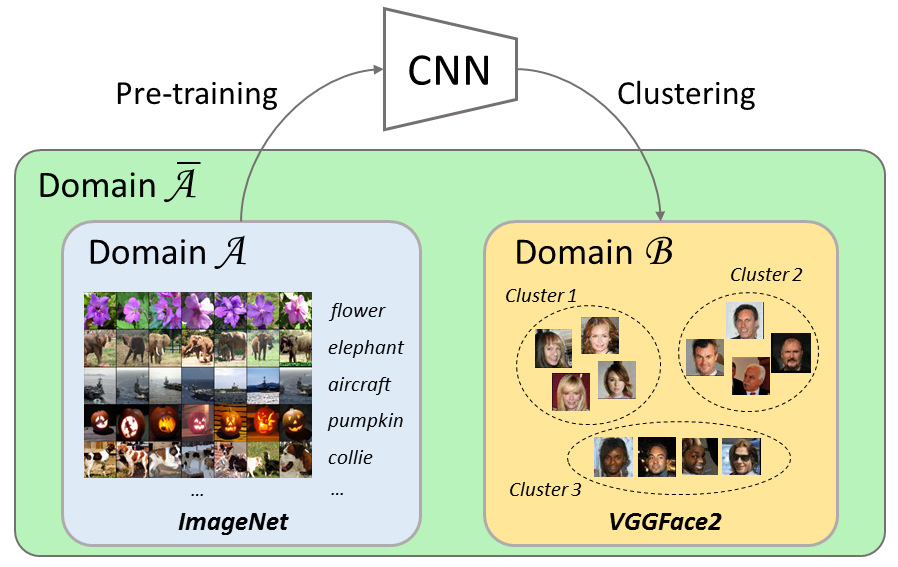}
\caption{Demonstration of the general discrimination ability: the features of a model pre-trained on ImageNet could be effectively used to cluster faces from VGGFace2~\cite{vggface2} by their genders and races. It indicates that to some extent, a CNN model pre-trained on domain $\mathcal{A}$ also has the ability to discriminate the patterns belonging to domain $\mathcal{\bar{A}}$, although it has never been trained to do so.}
\vspace{-1em}
\label{fig:intro}
\end{figure}

\begin{figure}[t]
\centering
\includegraphics[width=0.48\textwidth]{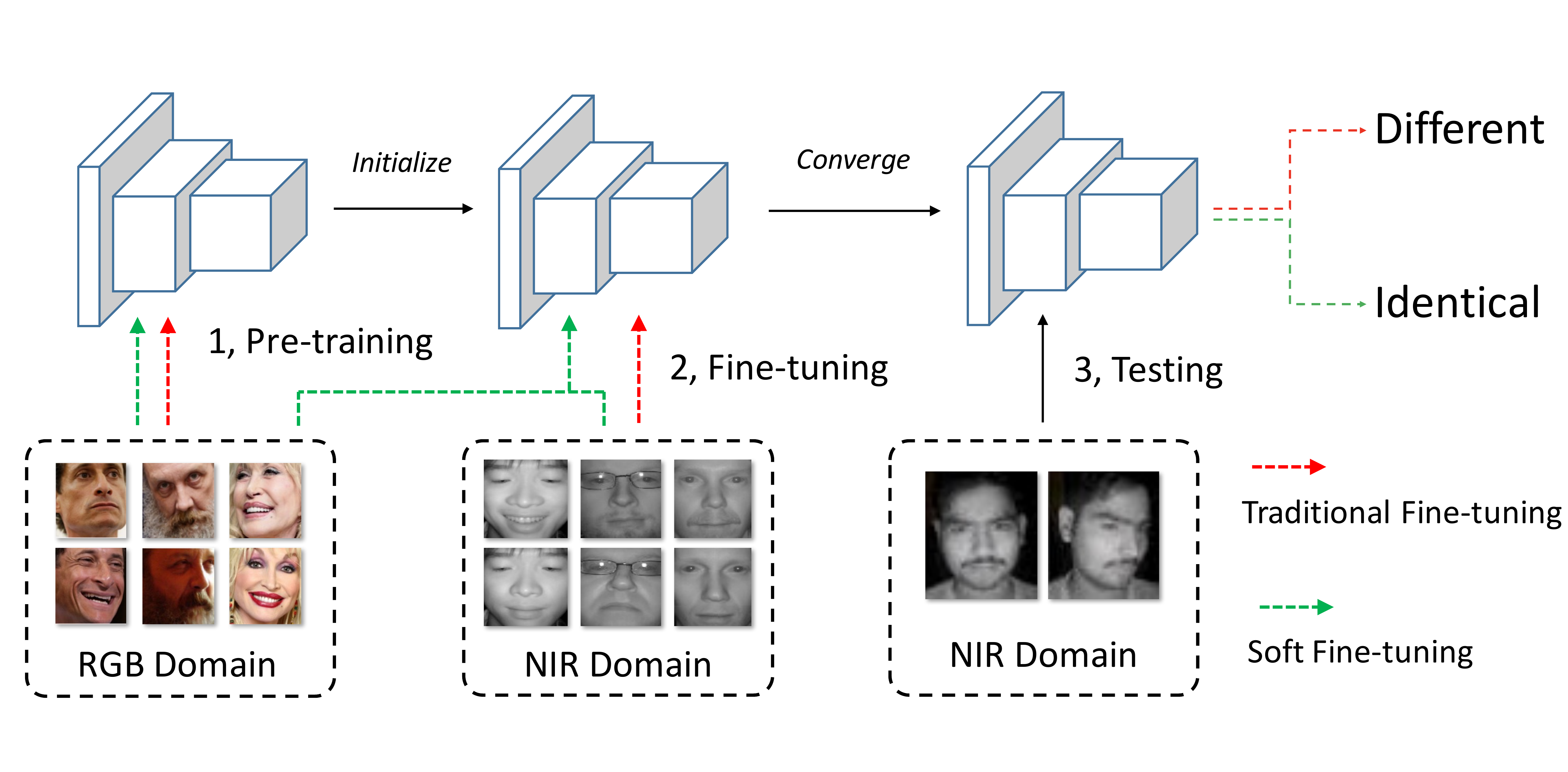}
\vspace{-1em}
\caption{Comparison between traditional fine-tuning and soft fine-tuning. By involving source dataset into soft fine-tuning, the network is supposed to be more robust to the bias in small-scale target dataset.}
\vspace{-1em}
\label{fig:insp3}
\end{figure}

Despite previous success, the existing fine-tuning technique still has several drawbacks when applied on a small target dataset. Firstly, the last layer of the network is replaced and randomly re-initialized, which means in the beginning of the fine-tuning, the gradients generated by the randomly initialized layer will be somehow noisy for the pre-trained parameters. Such noisy gradients may mislead the back-propagation of the shallow layers and thus slower the convergence. Secondly, as the network is trained only using the target dataset, the network may tend to learn the bias of the data distribution especially when the target dataset is too small. In Fig.~\ref{fig:insp3} we take face recognition task as an example. In this scenario, the source domain refers to general RGB images crawled from Internet, and the target domain refers to near-infrared-ray (NIR) images of human faces. Considering the difficulties in data collection, it is reasonable to assume that the source dataset contains millions of faces in large variance (\eg, with various poses and expressions) while the target dataset contains only hundred of human faces in small variance (\eg, with various expression but all frontal faces). As a result, the fine-tuned model will probably degrade its robustness to pose variation and lead to more failure cases in NIR domain since it tends to learn the bias of the small target dataset.

With the observations above, naturally we are wondering whether it is possible to take more advantages of the source domain in the fine-tuning process? And how? To the first question our answer is yes. Given a CNN model pre-trained on domain $\mathcal{A}$, we argue that the model not only has the ability to discriminate the patterns belonging to $\mathcal{A}$, to some extent it also has the ability to discriminate the patterns belonging to $\mathcal{\bar{A}}$ which it has not been trained with. We call the ability {\it general discrimination}. Fig.~\ref{fig:intro} provides an example to demonstrate the general discrimination ability: a CNN model is pre-trained for image categorization task on ImageNet~\cite{imagenet} (denoted by domain $\mathcal{A}$) while it has never been supervised to discriminate the genders by human faces (denoted by domain $\mathcal{B} \subset \mathcal{\bar{A}}$). However, when we use the fully-connected (fc) layer of the CNN model as features and perform clustering, the human faces are automatically clustered by their genders in an unsupervised manner. Moreover, it implies that the discriminative ability of a CNN model on domain $\mathcal{B}$ could be potentially enhanced by its general discrimination ability obtained from domain $\mathcal{A}$. Under the assumption, we suggest that the source dataset should not be simply discarded when transferring a pre-trained CNN model to the target domain; on the contrary, the source dataset should be involved into the fine-tuning process to keep the general discrimination ability of a CNN model.

Now we answer the second question, \ie, how to better utilize the knowledge from the source domain, in this paper we propose a novel transfer learning technique, namely {\it soft fine-tuning}. In the beginning of soft fine-tuning, instead of replacing the last classifier layer pre-trained on source domain, we add a new classifier layer for the target domain task while keeping the original layer as well. Then the training samples from both source and target datasets are fed to the network and the losses of two domains are optimized jointly. Then with the fine-tuning process going, the weight on the loss of the source domain is gradually decreased to zero, and finally only the loss of the target domain remains. By doing so, the network will be focused on the target task in the home stretch.

Compared to traditional fine-tuning, the advantages of the soft fine-tuning technique are twofold: 1) at early steps, the pre-trained last layer as well as the training data from source domain will continuously contribute a stable gradient to the shallow layers, which offsets some negative effects of noisy gradient generated by the new-added layer. It will make the fine-tuning process converge faster; 2) the network is required to keep its general discrimination ability learned from source domain, which prevents the model from overfitting the bias of target dataset and hence improves the total discriminative ability as well as robustness of the model, especially for the target datasets in small scales.

The main contributions of the paper are summarized as:
\begin{itemize}
\setlength{\itemsep}{0pt}
\setlength{\parskip}{0pt}
\setlength{\parsep}{0pt}
\vspace{-0.05in}
  \item First, the concept of general discrimination ability is introduced in transfer learning. Experimentally we show that in the fine-tuning process, the general discrimination ability of a network learned from the source domain will enhance its total discriminative ability on the target domain.
      \vspace{0.02in}
  \item Next, a novel transfer learning technique, \ie soft fine-tuning, is proposed to better utilize the knowledge from the source domain. The soft fine-tuning technique is independent of network architecture, which means it could be easily adopted for various models and tasks.
      \vspace{0.02in}
  \item Finally, we conduct experiments on various recognition tasks: action recognition~\cite{Yao12}, fine-grained recognition~\cite{dogs,aircraft} and NIR face recognition~\cite{oulu}. Our results outperform the state-of-the-arts significantly, highlighting the effectiveness of the soft fine-tuning technique.
\end{itemize}

\section{Related Work}
\label{sec:related}
\noindent{\textbf{Transfer learning.}} Due to the difficulties in data collection and labeling in some specific tasks, transfer learning, whose objective is to transfer the knowledge learned from a general source domain to a specific target domain, has been regarded as an effective way to solve the problem and hence attracted extensive research interests~\cite{ZhouLKO018,rcnn,inat,nat}. While the earlier works~\cite{Razavian2014CNNFO,donahue14} directly used the pre-trained network as a feature extractor and applied a simple classifier (\eg SVM) to specific tasks, fine-tuning the pre-trained network by the samples from target domain becomes a standard approach in most transfer learning works~\cite{rcnn,NIPS2014_5347,nat} nowadays. In this way, the fine-tuned networks are supposed to have better discriminative ability on the target domain. 

Recently, there are some works trying to provide a better understanding on the fine-tuning process and further improve it. 
Some works propose novel methods of reusing layers trained on the ImageNet dataset to compute mid-level image representation~\cite{Oquab14}.
Other works, from view of data, study the relationship of transfer learning and dataset. 
\cite{Huh2016WhatMI} studies the factors of feature learning and suggests using more data per class
in transfer learning. Sun \etal.~\cite{jft} propose a larger dataset JFT-300M and improve many vision tasks.
Recently, some works try to mine the connection between transfer learning and domain similarity. Azizpour \etal. ~\cite{RSMC16} implement detailed experiments on list of transfer learning tasks, and measure the similarity of each task with the original ImageNet~\cite{imagenet}. Cui \etal.~\cite{nat} search similarity categories from both ImageNet~\cite{imagenet} and iNaturalist~\cite{inat} datasets to improve fine-grained categorization. The difference of our soft fine-tuning and existing transfer learning approaches is two-fold: on one hand, soft fine-tuning does not require additional datasets, the performance can be improved by single source domain. On the other hand, it is independent of network architecture, and could be easily adopted for various models and tasks.

\noindent{\textbf{Single image action recognition.}} There are two popular strategies for single image action recognition: context-based approaches and part-based approaches. Context-based approaches try to capture interacting objects cues, and always requires object proposals or detectors~\cite{contextualaction,mini}. Part-based approaches focus on human parts. A simple approach can be combining global appearance and part appearance, and concatenating their features to form the representations~\cite{SMSP}. Zhao \etal.~\cite{Zhao_2017_ICCV} define actions on part level and propose Part Action Network that learns mappings from part appearance to part actions. In this paper, this task is mainly for ablation study.

\noindent{\textbf{Infrared face recognition.}} Despite the rapid development on visible light (VIS) face recognition, the tasks on invisible light domain remains a challenging problem. In surveillance scenarios, Near Infrared Ray (NIR) images are important to track identities and their actions. Researchers collect Oulu-CASIA NIR\&VIS dataset~\cite{oulu} and CASIA NIR-VIS 2.0 face dataset~\cite{casia} to evaluate approaches of cross spectral recognition. Recently, ~\cite{notafraid} hallucinates a VIS image from NIR sample, and extracting low-rank embedding of DNN features on such outputs. He \etal.~\cite{wassercnn} minimize wasserstein distance of NIR and VIS distributions, and transfer knowledge learned from VIS domain to NIR domain. In this paper, we take VIS as source domain and NIR as target domain. So our target is to reach best performance on NIR-NIR verification. Since there is few NIR-NIR public dataset, we employ cross spectral dataset and propose a new protocol on Oulu-CASIA dataset and evaluate our method. The results on Oulu CASIA dataset demonstrate that our method is less affected by bias.

\noindent{\textbf{Fine-grained recognition.}} The fine-grained tasks focus on distinguishing fine-grained categories or subcategories like subspecies of dogs~\cite{dogs} or foods~\cite{food101}. Feature coding approaches perform promising results on mining local features. The second order bilinear features are shown to be effective by B-CNN~\cite{BCNN}. Since discriminative features of fine-grained categories locally distribute, attention-based approaches provide promising results~\cite{lookingcloser}. ~\cite{multitaskdomain,noisybird} collect additional web images to augment the datasets lacking of training samples. Our method improves fine-grained classification by keeping general discrimination, without additional data, similar categories or feature coding approaches. In this task we show state-of-the-art results on small-scale datasets.

\section{Soft Fine-tuning for Visual Recognition}
\label{sec:method}
\begin{figure*}
\centering
\includegraphics[width=0.8\textwidth]{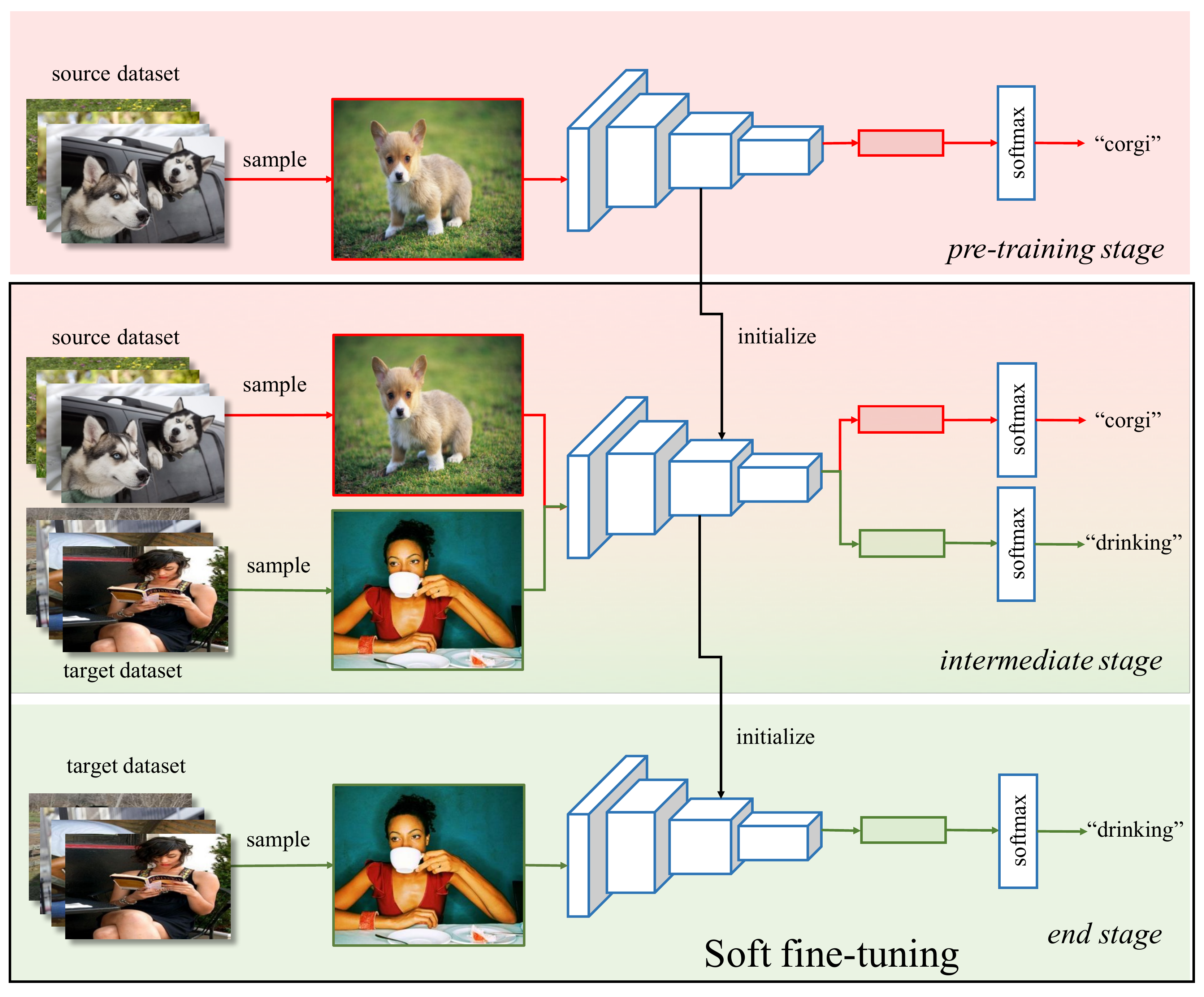}
\caption{Framework of our soft fine-tuning. Red blocks/lines refer to images and features of the source domain, green refers to the target domain.
The hybrid region is an intermediate stage.  It takes pre-trained model as initialization, samples from source dataset and target dataset simultaneously, and optimize the network for both tasks. Finally, the network are optimized only on target task for best performance. The whole process of intermediate stage and the end stage is the proposed soft fine-tuning.}
\label{fig:pipeline}
\end{figure*}

\begin{table*}
\small
\begin{center}
\begin{tabular}{|c|c|c|c|c|c|c|}
\hline
Dataset & target domain & tasks & \# categories & \# training samples & \# validation samples & evaluation metric\\
\hline
ILSVRC 2012~\cite{imagenet} &  & object & 1000 & 1,281,167 &  50,000 & -\\
Stanford 40~\cite{Yao12} & \checkmark & action & 40 & 4,000 &  5,532 & mAP\\
Stanford Dogs~\cite{dogs} & \checkmark & fine-grained & 120 & 12,000 & 8,580 & accuracy\\
Aircraft~\cite{aircraft} & \checkmark & fine-grained & 100 & 6,667 & 3,333 & mean accuracy\\
VGGFace2~\cite{vggface2} &  & face & 9,131 & 3.14M & 0.17M & -\\
Oulu CASIA~\cite{oulu} & \checkmark & face & 80 & 24,288 & 3,840 & TAR@FAR\\
\hline
\end{tabular}
\end{center}
\caption{Statistic of source and target datasets used in this paper.}
\vspace{-1em}
\label{statistic}
\end{table*}

\begin{table*}
\footnotesize
\begin{center}
\begin{tabular}{|c|c|c|c|c|c|}
\hline
tuning method & Network & ImageNet val. Top-1 acc. & w/o ft & mAP & gain\\
\hline
pre-training & MobileNetV2  & 71.3 & 73.1 & - & -\\
fine-tuning & MobileNetV2  & - & - & 80.9 & -\\
soft fine-tuning & MobileNetV2 & - & - & \textbf{84.0} & +3.1\%\\
\hline
pre-training & ResNet-50 & 74.9 & 80.2 & - & -\\
fine-tuning & ResNet-50 & - & - & 84.8 & -\\
intermediate stage & ResNet-50 & - & - & 87.1 & +2.3\%\\
soft fine-tuning & ResNet-50 & - & - & \textbf{88.5} & +3.7\%\\
\hline
pre-training & InceptionV4 & 80.0 & 86.8 & - & -\\
fine-tuning & InceptionV4 & - & - & 92.2 & -\\
soft fine-tuning & InceptionV4  & -& -& \textbf{93.2} & +1.0\%\\
\hline
\end{tabular}
\end{center}
\caption{Soft fine-tuning vs. fine-tuning.}
\vspace{-1em}
\label{ablation}
\end{table*}
In this section we conduct extensive studies on general discrimination and propose our method of soft fine-tuning.
The section is organized as follows: in Sec.\ref{3p1} we demonstrate what is general discrimination and its importance in transfer learning. The method of keeping it is stated in Sec.\ref{3p2}. By adjusting weights of source and target domain loss functions, we introduce the proposed soft fine-tuning in Sec.\ref{3p3}, and
Sec.\ref{3p4} figures the key factor of yielding better general discrimination.
All the studies in this section are implemented with the following settings:
we use three models of different sizes: MobileNetV2 (14M,~\cite{mobilenetv2}), ResNet-50 (98M,~\cite{resnet}) and InceptionV4 (164M,~\cite{inceptionv4}).
Models are pre-trained on ImageNet~\cite{imagenet} and used in Stanford-40~\cite{Yao12}. 
The Stanford-40~\cite{Yao12} dataset contains $40$ categories on human actions in still images, it is chosen since there are few categories about actions in ILSVRC 2012~\cite{ilsvrc} (more details can be found in Table.\ref{statistic}).
We choose ``baseline network" in~\cite{Zhao_2017_ICCV} in the experiment,
which receives both the whole image and the bounding box image, concatenates their features and obtains classification results.
In training stage, images are resized to 256/256/320 and randomly cropped to $224/224/299$ for MobileNetV2/ResNet-50/InceptionV4.

\subsection{Exploring General Discrimination}\label{3p1}
As defined in Sec.\ref{sec:intro}, general discrimination describes the ability of a network on distinguishing unknown patterns.
An evidence indicating the existence of general discrimination of is that just use a pre-trained network as feature extractor and trains classifiers to achieve well performance~\cite{donahue14,ZhouLKO018}.

In Table.\ref{ablation}, we explore the existence of general discrimination on a representative action recognition dataset.
The {\it w/o ft} tag means without fine-tuning and SVMs are used as classifiers. 
According to Table.\ref{ablation}, models pre-trained on large source domain can provide well results on unknown categories (from 73.1\% to 86.8\%). 
Another observation is that models with better general discrimination also performs better in fine-tuning. 

\noindent \textbf{Conclusion:}
1) Models trained on large-scale source domain have general discrimination, which is embedded in features.
2) General discrimination is related to the performance of transfer learning.





\subsection{Preserving General Discrimination in Transfer Learning}\label{3p2}
Now imagine a fine-tuning process: It transfers a network from a large source domain to a small target domain. 
When the fine-tuning begins, the network is required to learn knowledge on the target domain, and
its target-specific discrimination improves. 
However, if the scale of the target dataset is small, general discrimination of the network degrades.

By reviewing the fine-tuning stage, the reason that target-specific discrimination can be improved is because of the restriction of training loss. However, since the network is not optimized on the source domain, the general discrimination degrades. 
According to the observation above that general discrimination is related to transfer learning performance, 
if we can keep the general discrimination as well, the combination of both discrimination may be able to benefit performance! Based on this idea, we propose an intermediate stage (see Fig.\ref{fig:pipeline}). In this stage we optimize the network by two loss functions: the source domain loss and the target domain loss. In each batch we sample two images from both domains, and feed forward them to the same network.  Features of them are separately classified by source and target domain classifiers, and the network receives gradients from both loss functions.


We verify whether keeping general discrimination improves transfer learning in Table.\ref{ablation}. See the comparison of ``fine-tuning" and ``intermediate stage" with ResNet-50, the latter model provides better results (84.8\% - 87.1\%).

\noindent \textbf{Conclusion:} The total discriminative ability of a network on the target domain gets enhanced by preserving its general discrimination.

\subsection{When Do We Need General Discrimintation?}\label{3p3}
We can rethink the source domain loss as well. It has its own risks and benefits: the source domain loss helps keep general discrimination, however, it's gradients may be different, or even opposite with gradients of target domain loss, which may limit its performance.
It inspires us to adapt weights of two loss functions.
Considering that target-specific discrimination is directly related to the accuracy, we gradually 
decay the source domain loss as follows:

\begin{equation}
loss = (1-\alpha)  loss_{src} + loss_{tar}
\label{eq1}
\end{equation}
where 
\begin{equation}
\alpha = min(1,num_{epoch}/E).
\label{eq2}
\end{equation}
$num_{epoch}$ refers to the epoch index, $E$ is a scale coefficient.

Since we ``softly" transfer from the source domain to the target domain,
we name this learning method as ``soft fine-tuning". The ``intermediate stage" is essentially soft fine-tuning with $\alpha = 0$.
Note we stop training soon after $\alpha$ reaches 1, otherwise the network may lose general discrimination again.

To evaluate the necessity of adjusting the weight of the source domain loss, we compare settings of $\alpha=0$ and using Eqn.\ref{eq1} for ResNet-50 in Table.\ref{ablation}.
By reducing the source domain loss to zero, the network performs better and reaches mAP of 88.5\%. 
Moreover, we generalize the study on various models.
We observe improvement of +3.1\%, +3.7\% and +1.0\% on MobileNetV2~\cite{mobilenetv2}, ResNet-50~\cite{resnet} and InceptionV4~\cite{inceptionv4} respectively. The results demonstrate that soft fine-tuning can provide better results consistently with models of various sizes.

The proposed soft fine-tuning has the following advantages: 
\begin{itemize}
\vspace{0.02in}
\item Soft fine-tuning accelerates the convergence. In typical fine-tuning method, the last layer of models trained on ILSVRC 2012~\cite{ilsvrc} (which has $1000$ activations) is replaced by a target-specific classifier. Since the classifier is randomly initialized, it propagates noisy gradients to shallow layers, and slowers the training convergence 
($\partial L / \partial w_{conv}= \partial L / \partial x_{fc} * \partial x_{fc} / \partial w_{conv}$, where $\partial L / \partial x_{fc}$ is related to $w_{fc}$). In soft fine-tuning, shallow layers receive gradients from both source and target domain loss. The source domain loss propagates qualified gradients and thus it ``rectifies" the whole gradients. The shallow layers receive moderate gradients and training convergence is accelerated. We verify this in Sec.\ref{stateofart}.
\vspace{0.02in}
\item Soft fine-tuning improves the performance of transfer learning. By keeping training on the source domain, it holds general discrimination. Some knowledge may not be learned from the target domain training data because of bias, however, it is probably captured in large source dataset. In Sec.\ref{infared} we discuss this.

\end{itemize}

\noindent \textbf{Conclusion:} We propose a novel soft fine-tuning approach for transfer learning. By decay the loss on the source domain, it better utilizes the knowledge learned from the source domain and accelerates the convergence.

\begin{table}
\footnotesize
\begin{center}
\begin{tabular}{|c|c|c|}
\hline
source domain & target domain & mAP\\
\hline
ImageNet & Stanford 40 & 88.5\\
ImageNet & Stanford 40 \& Stanford Dogs & 88.2\\
ImageNet (10\% categories) & Stanford 40 & 88.0\\
ImageNet (10\% images) & Stanford 40 & 87.2\\
\hline
\end{tabular}
\end{center}
\caption{What brings general discrimination?}
\vspace{-1em}
\label{sourcerequire}
\end{table}

\subsection{What Brings General Discrimination?}\label{3p4}
Despite the benefit of general discrimination in transfer learning, we wonder where is general discrimination from? In previous sections we mention that the source domain is a large dataset. Here we explore the hidden reasons.
We make a comparison in Table.\ref{sourcerequire}. The baseline is using ResNet-50 and soft fine-tuning on Stanford 40 dataset~\cite{Yao12}.

First, we measure if a small-scale dataset, or a multi-task learning framework helps. We use a pre-trained model and fine-tune it simultaneously on both Stanford Dogs~\cite{dogs} and Stanford-40~\cite{Yao12} datasets. The Stanford Dogs~\cite{dogs} dataset is another small dataset, whose statistic is shown in Table.\ref{statistic}.
In this trial, the loss function of Eqn.\ref{eq1} can be rewritten as:
\begin{equation}
loss = (1-\alpha)  loss_{src} + loss_{tar} + loss_{tar'}
\label{eq3}
\end{equation}
where $tar$ and $tar'$ refer to the two target datasets. Compared with result of training on a single target domain, it yields even worse result.
Such observation implies two points: 1) another small-scale dataset provides no additional general discrimination. 2) a multi-task learning framework, which requires sharing knowledge among branches, cannot provide general discrimination as well. The reason of worse result may be the introducing of bias on Stanford Dogs~\cite{dogs}.

It leads to the following question: is more training data or more categories the key factor of general discrimination? We design more experiments to answer the question. As shown in Table.\ref{sourcerequire}, we replace the source domain by two variants of ILSVRC 2012~\cite{ilsvrc}. One remains randomly chosen 10\% of categories, in which all images are preserved. The other preserves 10\% of images for each category.
The image-preserved trial performs better than the category-preserved trail (88.0\% vs 87.2\%). The reason may be that abundant images have large intra-class variance and mapping such samples helps recognize patterns. In ~\cite{nat,Huh2016WhatMI} the authors show similar observations that in some tasks using less categories provides even better performance.

\noindent \textbf{Conclusion:} Larger dataset provides better general discrimination, which mainly comes from abundant samples instead of increasing of categories.

\section{Experiments}
\label{sec:experiment}
\subsection{Experiments Setup}
In this section we evaluate detailed performance of soft fine-tuning as well as convergence speed.
We evaluate our approach on three tasks: action recognition, fine-grained recognition and NIR face recognition:
\begin{itemize}
\item For action recognition we use Stanford-40 dataset ~\cite{Yao12} as the target domain, and ILSVRC 2012~\cite{ilsvrc} as the source domain. The Stanford-40 dataset contains 40 categories and uses 4000 images for training. In this dataset humans are always interacting with objects or scenes, such as ``climbing", ``riding a horse", ``texting message" and so on. The evaluation metric is mean average precision (mAP).

\item We evaluate the soft fine-tuning on two fine-grained recognition
datasets: Stanford Dogs~\cite{dogs} and Aircraft~\cite{aircraft}. The former dataset collects $120$ subspecies such as ``Chihuiahua", ``Papillon", ``Beagle" and so on, while Aircraft~\cite{aircraft} focus on variants such as ``Boeing 737-300" and ``Boeing 737-400". The statistics of these datasets can be found in Table~\ref{statistic}. ILSVRC 2012~\cite{ilsvrc} is chosen as the source domain.

\item In face recognition, the source domain is VIS (RGB) domain, and target domain is NIR domain. We use VGGFace2~\cite{vggface2} as RGB dataset, and Oulu CASIA~\cite{oulu} as NIR dataset. 
The Oulu CASIA dataset~\cite{oulu} collect both VIS and NIR images for $80$ identities, with $3$ light environments and $6$ expressions.
Since it is proposed for cross spectral recognition, we define a protocol in this paper to evaluate NIR-NIR 1:1 verification performance as follows: we use 20 identities (P001-P020) as test set and the others (P021-P060) as training data. In the test phase, two images generate one pair. We calculate cosine similarity of each pair as used in~\cite{vggface2} to measure whether the pair refers to the same identity. 
The evaluation metric is true accept rates (TAR) given false accept rates (FAR).
Such settings simulate the real-world applications where one can train models on limited identities and have to apply it to strangers.
\end{itemize}

In our experiments we set batch-size to $32$ for all tasks. The learning rate is $0.001$ for fine-grained recognition and $0.0001$ for others. We train $100/25/5$ epochs for action/fine-grained/face recognition, and the corresponding $E$ is $60/20/5$.
To mitigate overfitting on Aircraft~\cite{aircraft}, we use the following augmentation:
scale and aspect ration variation, color noise and scale jittering. Label smoothing (LSR, ~\cite{inceptionv3}) is also used.

We implement our approach with PyTorch\footnote{https://github.com/pytorch}, codes and models will be released.







\subsection{Action Recognition}

\begin{table}
\small
\begin{center}
\begin{tabular}{|c|c|c|}
\hline
Method & Network & mAP \\
\hline
Top-down pyramid~\cite{topdown}  & VGG-16 & 80.6\\
ActionMask~\cite{mini}  & VGG-16 & 82.6\\
Yang \etal.~\cite{pbnc}  & VGG-19 & 86.9\\
R*CNN~\cite{contextualaction}  & VGG-16 & 90.9\\
Part Action Network~\cite{Zhao_2017_ICCV}  & ResNet-50 & \textbf{91.2}\\
\hline
ours &  ResNet-50 & \textbf{92.2}\\
\hline
\end{tabular}
\end{center}
\caption{Performance Comparison on Stanford-40 dataset.}
\vspace{-1em}
\label{actionres}
\end{table}

\begin{figure}
\centering
\includegraphics[width=0.4\textwidth]{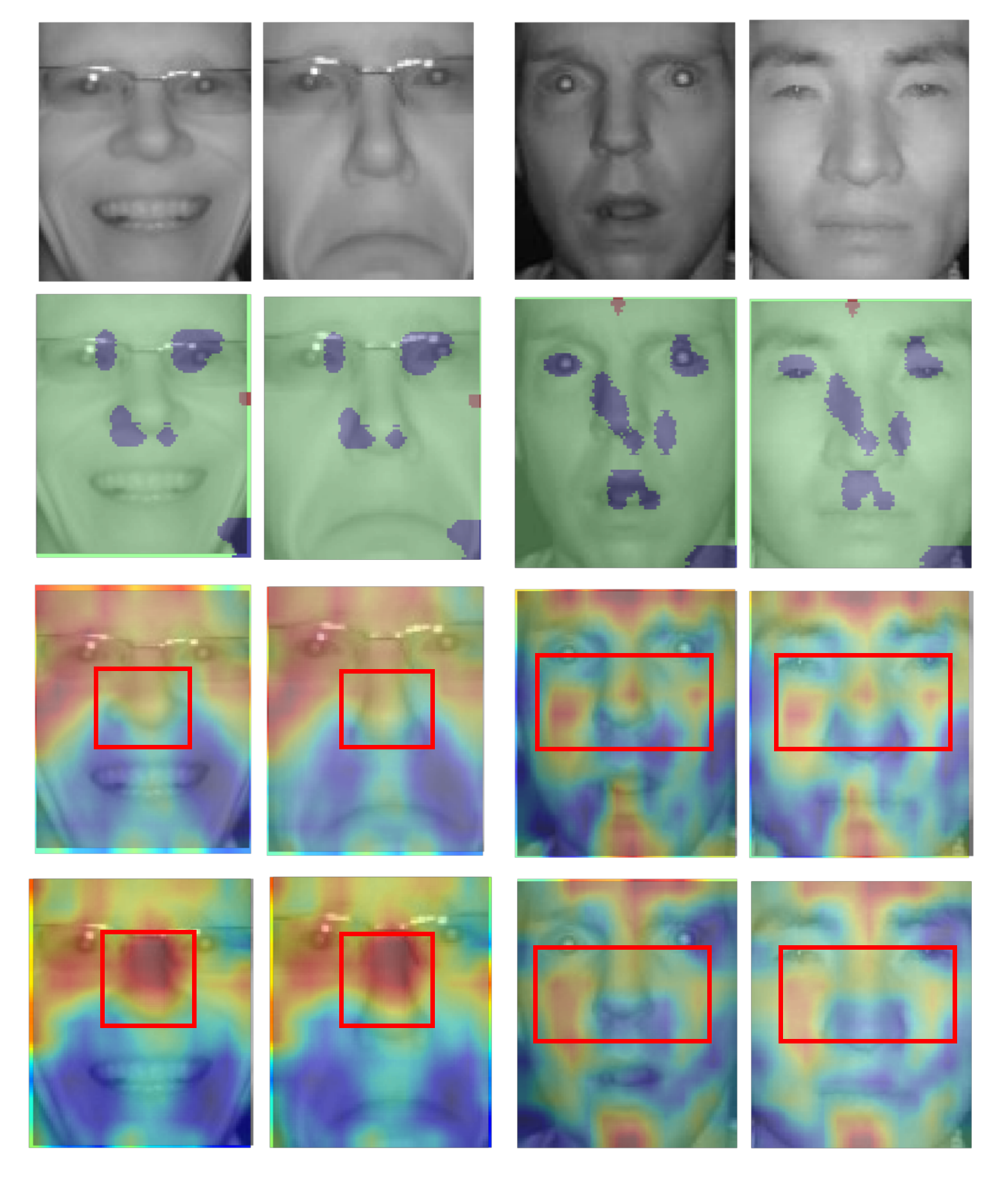}
\caption{Visualization of feature map response. The first two faces belong to the same identity. We visualize input images, responses of pre-trained model, fine-tuned model and softly fine-tuned model from top to bottom.
For faces cosine similarity is calculated by each region. Difference is marked by red boxes. See text for details.}
\vspace{-1em}
\label{fig:vis}
\end{figure}

\begin{table*}
\small
\begin{center}
\begin{tabular}{|c|c|c|c|c|}
\hline
\multirow{2}{*}{Method} & \multicolumn{4}{|c|}{1:1 verification TAR (\%)}\\
\cline{2-5}
& FAR@1e-2 & FAR@1e-3 & FAR@1e-4 & FAR@1e-5\\
\hline
pre-training & 22.3 & 10.7 & 6.3 & 4.5 \\ 
fine-tuning & 80.4 & 64.5 & 47.4 & 40.1 \\
soft fine-tuning & 86.8 & 76.1 & 66.7 & 58.7\\
fine-tuning (trained on 2 expressions) & 67.8 & 54.2 & 44.9 & 37.9 \\
soft fine-tuning (trained on 2 expressions) & 84.2 & 71.3 & 58.7 & 48.6\\
\hline
\end{tabular}
\end{center}
\vspace{-1em}
\caption{Performance Comparison (TAR) on NIR face recognition. Soft fine-tuning trained on less expressions (line 5) performs better than fine-tuning trained on full expressions (line 2).}

\label{face}
\end{table*}

\begin{table*}
\small
\begin{center}
\begin{tabular}{|c|c|c|c|c|c|}
\hline
Method & source dataset & Input Size & network & Stanford Dogs & Aircraft\\
\hline
Bilinear-CNN~\cite{BCNN} & ImageNet & $448\times448$ & VGG-19 & - & 84.1\\
Zhang \etal.~\cite{pickingfilter}   & ImageNet & $224\times224$ & VGG & 72.0 & -\\
RA-CNN~\cite{lookingcloser} & ImageNet & $448\times448$ & VGG-19/16 & 87.3 & -\\ 
DLA~\cite{DLA} & ImageNet & $448\times448$ & VGG & - & \textbf{92.6}\\ 
Cui \etal.~\cite{nat} & ImageNet\&iNat & $299\times299$ & InceptionV3 & 85.2 & 86.1\\
Cui \etal.~\cite{nat} & ImageNet\&iNat & $448\times448$ & Inception-ResNetV2 SE & \textbf{88.0} & 90.7 \\
\hline
fine-tuning & ImageNet & $299\times299$ & InceptionV4& 84.7 & 87.1\\
soft fine-tuning (ours)  & ImageNet & $299\times299$ & InceptionV4& 91.0 & 88.4\\
soft fine-tuning (ours)  & ImageNet & $448\times448$ & InceptionV4& \textbf{91.7} & 91.1\\
\hline
\end{tabular}
\end{center}
\caption{Performance Comparison on fine-grained dataset.}
\vspace{-1em}
\label{fineres}
\end{table*}

\begin{figure}[t]
\includegraphics[width=0.45\textwidth]{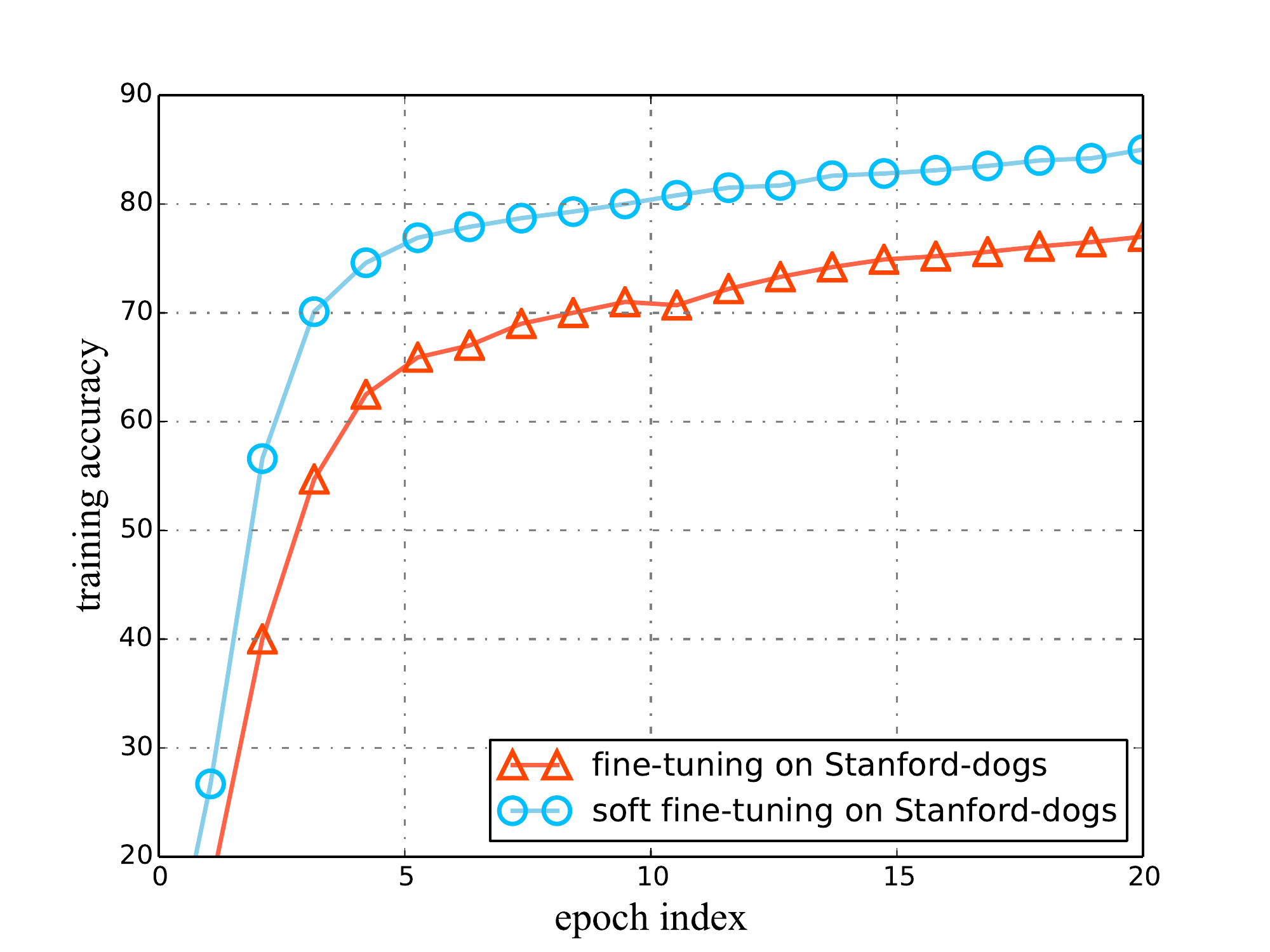}
\caption{Comparison of training speed  between soft fine-tuning and fine-tuning. 
The soft fine-tuning leads all the time, and keeps significant gain of accuracy.}
\vspace{-1em}
\label{fig:training}
\end{figure}

In the Stanford 40 dataset~\cite{Yao12}, we re-implement the Part Action Network~\cite{Zhao_2017_ICCV} and train it by soft fine-tuning. For fair comparison we only use ResNet-50 as in~\cite{Zhao_2017_ICCV}.
Note the part action network feeds $9$ images in a sample: bounding box image, the whole image and other 7 part images for head, torso, legs, arms and hands. In the soft fine-tuning framework, we use $1$ source domain image  and the $9$ images above to form a sample. Then samples are aggregated to batches.

We show the results in Table.\ref{actionres}. Zhao \etal.~\cite{topdown} learn some semantic detectors, and arrange semantic parts in top-down order, obtaining larger inter-class variance.
R*CNN~\cite{contextualaction} is proposed to capture interactive objects and reaches mAP of 90.9\%. The method of Part Action Network~\cite{Zhao_2017_ICCV} defines 7 parts and provides part-level action annotations. The network first predict part actions, and then uses features of part actions,
human appearance and context to classify a sample.

Our method based on soft fine-tuning outperforms the state-of-the-art and achieves mAP of 92.2\%. The improvement mainly comes from categories like ``smoking" (+5.0\%), ``taking photos" (+4.3\%) and ``texting message" (+2.2\%), where interacting objects are critical. In these scenario, general discrimination can be understood as the knowledge of various objects.

\subsection{Infrared Face Recognition}\label{infared}

We use a ResNet-50 network~\cite{resnet} pre-trained on VGGFace2~\cite{vggface2} for this task. In the training phase, both source and target tasks are modeled as classification (i.e. each identity is one category).
In the test phase, we remove the last 8631-way and 60-way fully connected layers (8631 and 60 refer to the amount of identities of the two domains), and use features of the second last layer. The similarity of two samples is calculated as cosine similarity of two features.

We report the results in Table.\ref{face}. Even without training on Oulu CASIA dataset~\cite{oulu}, a pre-trained model can rank samples based on its general discrimination.
By fine-tuning on P021-P060, TAR can be improved significantly in NIR domain. Our soft fine-tuning improves the performance by a large margin among all given FAR (+6.4\%, +11.6\%, +19.3\%, +18.6\%), especially on strict FAR.
When FAR = 1e-5, it even obtains about 50\% relative improvement. The practical meaning of such improvement is that few other identities can unlock one's NIR-based mobile phone, while the owner feels easier to unlock it.

We also implement experiments to measure whether soft fine-tuning helps on training bias. In Table.\ref{face}
we fine-tune/softly fine-tune models with only $2$ expressions (``surprise" and ``sadness") while test on all expressions. It results drop for both approaches, for fine-tuning, TAR drops more at loose FAR while soft fine-tuning drops more at strict FAR. Soft fine-tuning with less expressions even outperforms fine-tuning with full expressions consistently. It demonstrates that for fine-tuning hard negative samples obtain higher similarity, and soft 
fine-tuning is less affected by training bias.

In Fig.\ref{fig:vis} we visualize feature map response of \emph{conv4} layer of ResNet-50. Colors present cosine similarity of two features of the pair at the same position. On NIR face recognition, the method of soft fine-tuning yields higher similarity scores on noses when the same identity smiles (left), and smaller scores on the negative pair (right). This implies that our network trained by soft fine-tuning has knowledge on expressions, which is consistent with the numerical results in Table.\ref{face}.

\subsection{Fine-grained Recognition}\label{stateofart}
In this task we use the InceptionV4 network, which is a competitive model with Inception-ResNet-V2~\cite{inceptionv4} used in~\cite{nat}.
Like previous work~\cite{DLA,nat} we use no additional supervision (bounding box/part annotations), so
in one sample there are $2$ images coming from source domain and target domain respectively.

The results are shown in Table~\ref{fineres}, we list the source domain dataset, input size and backbone network for comprehensive comparison.
Bilinear-CNN~\cite{BCNN} represents feature of an image as a pooled outer product of activations derived from two CNNs. It essentially encodes high order representations, and achieves 84.1\% on Aircraft~\cite{aircraft}.
Zhang \etal.~\cite{pickingfilter} define some deep filters, and mine local discriminative features. RA-CNN~\cite{lookingcloser}, from the view of attention, generates patches for meaningful parts, and arranges features of parts to form comprehensive representations. Cui \etal.~\cite{nat} search similar categories among ImageNet and iNaturalist datasets, improving performance on many fine-grained tasks.

On the two fine-grained datasets, soft fine-tuning provides consistent gain compared with fine-tuning, especially on Stanford Dogs (+6.3\%).
Our approach outperforms the state-of-the-art significantly on Stanford Dogs dataset~\cite{dogs} (from 88.0\% to 91.0\%), with less data, smaller input size and competitive base model. With the input size increased, it reaches accuracy of 91.7\%. On Aircraft, it also obtains competitive results. 

We also record the training accuracy along with epochs in Fig.\ref{fig:training}. We find that soft fine-tuning leads fine-tuning method by at least $5$ epochs when accuracy $\ge$ 70\%. It also keeps about 7\% gain of accuracy all the time.
Such a result verifies the analysis in Sec.\ref{sec:method} that soft fine-tuning accelerates training convergence.

\section{Conclusion}
\label{sec:conclusion}
In this paper, we propose a novel and light-weighted framework of transfer learning: soft fine-tuning. We demonstrate that general discrimination is critical for the target domain, which mainly comes from abundant samples instead of categories. The method of soft fine-tuning keeps general discrimination and thus it uses better knowledge to predict. Our method outperforms the traditional fine-tuning method as well as the state-of-the-arts on various visual recognition tasks. As a simple but effective method which is independent of network architectures and types of tasks, we expect wide application of the soft fine-tuning technique in many other transfer learning tasks.

{\small
\bibliographystyle{ieee}
\bibliography{egbib}
}

\end{document}